\documentclass[letterpaper, 10 pt, conference]{ieeeconf}  

\usepackage{amsmath}
\usepackage{amssymb}
\usepackage{algorithm}
\usepackage{algorithmic}
\usepackage{graphicx}
\usepackage{microtype}  
\usepackage{booktabs}
\usepackage{multirow} 
\usepackage{cite,color}
\usepackage{subcaption}
\usepackage[colorlinks=true, linkcolor=lightblue, citecolor=lightblue, urlcolor=lightblue]{hyperref}
\definecolor{lightblue}{rgb}{0.0, 0.4, 0.8}

\IEEEoverridecommandlockouts                              

\title{\LARGE \bf
Robust Offline Imitation Learning Through State-level\\ Trajectory Stitching
}

\author{Shuze Wang$^{1, \dagger}$, Yunpeng Mei$^{1, \dagger}$, Hongjie Cao$^{1}$, Yetian Yuan$^{1}$, Gang Wang$^{1,*}$, Jian Sun$^{1}$, Jie Chen$^{2,1}$ 
\thanks{This work was supported in part by the National Key R\&D Program of China under Grant 2021YFB1714800 and the National Natural Science Foundation of China under Grants U23B2059, 62173034, 61925303, and 62088101.}
\thanks{$^{1}$ National Key Lab of Autonomous Intelligent Unmanned Systems, Beijing Institute of Technology, Beijing 100081, China}%
\thanks{$^{2}$ Department of Control Science and Engineering, Harbin Institute of Technology, Harbin 150001, China
}
\thanks{*Point of contact: gangwang@bit.edu.cn}
\thanks{$\dagger$ These authors contributed equally to this work.}
}

\begin{document}

\maketitle
\thispagestyle{empty}
\pagestyle{empty}


\begin{abstract}
Imitation learning (IL) has proven effective for enabling robots to acquire visuomotor skills through expert demonstrations. However, traditional IL methods are limited by their reliance on high-quality, often scarce, expert data, and suffer from covariate shift. To address these challenges, recent advances in offline IL have incorporated suboptimal, unlabeled datasets into the training. In this paper, we propose a novel approach to enhance policy learning from mixed-quality offline datasets by leveraging task-relevant trajectory fragments and rich environmental dynamics. Specifically, we introduce a state-based search framework that stitches state-action pairs from imperfect demonstrations, generating more diverse and informative training trajectories. Experimental results on standard IL benchmarks and real-world robotic tasks showcase that our proposed method significantly improves both generalization and performance. 
Videos are available on \href{https://youtube.com/playlist?list=PLUzaZQK4aUT3SER8YY-T_dBnE_ZS9_9_w&si=pThmzTtOQXbyA6Gj}{https://www.youtube.com/playlist?list=PLUzaZQK4aUT3SER8YY-T\_dBnE\_ZS9\_9\_w}.
\end{abstract}

\allowdisplaybreaks

\section{Introduction}
Imitation learning (IL) has quickly emerged as a promising framework for enabling robots to acquire complex tasks via expert demonstrations. Its potential spans real-world applications, such as industrial automation and household robotics, where it has been successfully applied for tasks like object manipulation and navigation \cite{BAKU, DiffusionPolicy, ACT3D,Aloha,RoboAgent, VLMIL, OCTO}. However, the effectiveness of IL is closely tied to the quality and diversity of expert data, which are often costly and time-consuming to collect, particularly for new or diverse tasks \cite{COFFLINEIL, DataQuality, LimitsIL}.


Manually collected data often suffer from behavioral inconsistencies due to varying operator skills, making it difficult to obtain reliable expert demonstrations. Moreover, these demonstrations may include suboptimal actions, such as retries of failed grasps \cite{GSR}. Directly applying behavioral cloning \cite{BC} to such data can lead to policies that replicate undesirable failure behaviors \cite{OfflineRLreview}. To overcome these limitations, algorithms must be capable of identifying and utilizing beneficial segments from imperfect demonstrations while filtering out certain irrelevant or noisy actions. This process, often referred to as trajectory stitching in reinforcement learning \cite{OfflineRLreview}, is crucial for improving the robustness of the policy learned from imperfect data.


\begin{figure}[t] \centerline{\includegraphics[width=84mm]{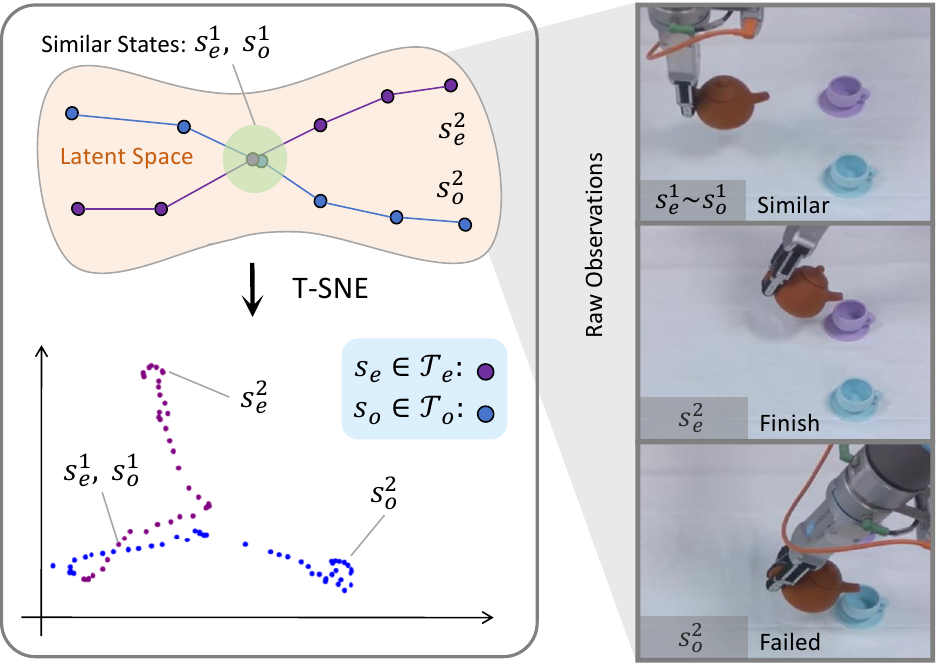}} \caption{Visualizing state embeddings of two trajectories using t-SNE projection, with expert and imperfect trajectories labeled in different colors.} \label{intro} 
\end{figure}

Recent research has addressed the problem of learning from mixed-quality datasets, where offline data include both expert data and suboptimal, unlabeled demonstrations \cite{BCND, DemoDICE, SMODICE, DWBC, UDS, ILID, SRA, GSR, SEABO}. These imperfect demonstrations often originate from failed attempts or random policies or even from different tasks executed within the same environment \cite{D4RL}. Despite not perfectly matching expert behavior, such data can provide valuable insights. For example, in the task of pouring water into the purple cup (see Figure \ref{intro}), a failed demonstration involving pouring water into the blue cup still shares crucial behavioral patterns with the expert’s actions in the early stages. Furthermore, imperfect trajectories may include states outside the expert distribution, which can offer rich transitional information from non-expert to expert states. These transitions teach the agent how to handle deviations from the expert's state distribution. However, such valuable information is often overlooked in many existing IL frameworks.

A limited number of works have explored leveraging imperfect demonstrations, adopting three main strategies: i) directly learning a policy from mixed-quality data, often via weighted behavioral cloning that prioritizes high-quality demonstrations \cite{DemoDICE, SMODICE, DWBC, DGIL}; ii) constructing reward functions and applying offline reinforcement learning methods \cite{UDS, SRA, SEABO}; and, iii) filtering datasets at the trajectory or state-action level before applying behavioral cloning techniques on the selected data \cite{ILID, BehaviorRetrieval}.


In this work, we focus on state-level filtering and selection, enabling the flexible stitching and retrieval of state segments of varying lengths. Our key insight is that imperfect data not only contains valuable trajectory fragments but also encodes rich environmental dynamics. Specifically, we propose state-based behavior retrieval (SBR), a novel approach for training policies on mixed-quality datasets. The first step involves learning a state similarity metric using both expert and imperfect demonstrations. To do this, we train a simplified world model on the union of expert and imperfect data to capture their shared environmental dynamics, thereby improving the robustness and temporal coherence of the learned latent representation. We then measure state similarity by computing the distances between states in the feature space \cite{BehaviorRetrieval, UVD}, as shown in Figure \ref{intro}. Using the t-distributed stochastic neighbor embedding (T-SNE) \cite{TSNE} algorithm, we visualize state embeddings of two trajectories. In the early phase, both trajectories involve the robot holding a teapot and moving it toward the teacup, with small distances between the state embeddings in the latent space. As the trajectories diverge, the latent space effectively captures this divergence, validating the efficacy of our approach. We introduce state-based behavior retrieval, which extracts positive behavioral patterns from imperfect data and stitches state-action pairs to create more informative trajectory segments that enhance data utilization for policy learning.

Our main contributions are summarized as follows: 
\begin{itemize} 
\item We propose an efficient positive behavior retrieval framework that combines a simplified world model with state-based search, enabling the effective exploitation of suboptimal data; 
\item We design an effective, lightweight offline IL algorithm based on the proposed framework; and, 
\item We evaluate our algorithm on standard IL benchmarks and real-world robotic tasks, showing superior performance in both generalization and policy learning. \end{itemize}

\section{Related Work}
\subsection{Offline Imitation Learning}
Offline IL focuses on learning policies from pre-collected datasets of demonstrations without further interaction with the environment. The most straightforward approach is behavioral cloning \cite{BC}, which minimizes the discrepancy between predicted and demonstrated actions. However, behavioral cloning suffers from covariate shift \cite{LimitsIL, RILSP}, leading to compounding errors during deployment. To mitigate this, various offline IL methods incorporate regularization techniques for the policy or use conservative value functions \cite{IQL,CQL}. Another prominent approach is inverse reinforcement learning \cite{IRL,MEIRL}, which aims to infer a reward function from expert demonstrations and iteratively optimize the policy. However, inverse reinforcement learning methods typically require substantial online interactions, which can reduce sample efficiency in offline settings \cite{AIRL,GAIL}. In contrast, our work assumes the availability of imperfect or low-quality data in the offline dataset and seeks to learn a robust policy without requiring any online interaction, thus addressing challenges specific to offline learning.

\subsection{Learning from Mixed-Quality Demonstrations}
Several methods have been proposed to tackle the challenge of learning from mixed-quality offline data \cite{DemoDICE, DWBC, UDS, ILID, OTIL, BehaviorRetrieval}. DemoDICE \cite{DemoDICE} enhances adversarial imitation learning by incorporating state-action distribution matching as a regularization term on the offline dataset. DWBC \cite{DWBC} combines expert and suboptimal trajectories, employing positive-unlabeled learning to develop a discriminator that extracts expert-like behaviors. OTIL \cite{OTIL} uses optimal transport theory to align unlabeled trajectories with expert trajectories by minimizing the Wasserstein distance. However, methods relying on a discriminator face limitations, as they only measure the similarity between expert and non-expert states. Behavior retrieval \cite{BehaviorRetrieval} extends this idea by employing an encoder to measure state similarity in an encoded space, effectively identifying beneficial behavior from suboptimal demonstrations. Our approach builds on these methods by extending the similarity measure to arbitrary states through a simplified world model and utilizing a state-based search process to retrieve more diverse, beneficial behaviors, resulting in more effective policy learning.

\subsection{Robotic Manipulation Policies}
Recent advancements in robotic manipulation have leveraged end-to-end deep learning architectures for controlling robotic arms \cite{Aloha, ACT3D, BAKU, DiffusionPolicy, VLMIL, OCTO, RoboAgent}. Methods such as diffusion policy \cite{DiffusionPolicy} frame action prediction as a denoising diffusion process, while Transformer-based models \cite{Aloha, ACT3D, BAKU, VLMIL, OCTO, RoboAgent} have demonstrated success in robotic policy learning. However, these approaches typically require large amounts of precise human demonstrations, which makes data collection both expensive and cumbersome. Our approach reduces the dependence on high-quality data by enabling the use of mixed-quality demonstrations, significantly alleviating the data collection burden while still achieving high performance.

\section{Policy Learning via State-based Search and Trajectory
Stitching}

\begin{figure*}[t] \centering \includegraphics[width=0.98\textwidth]{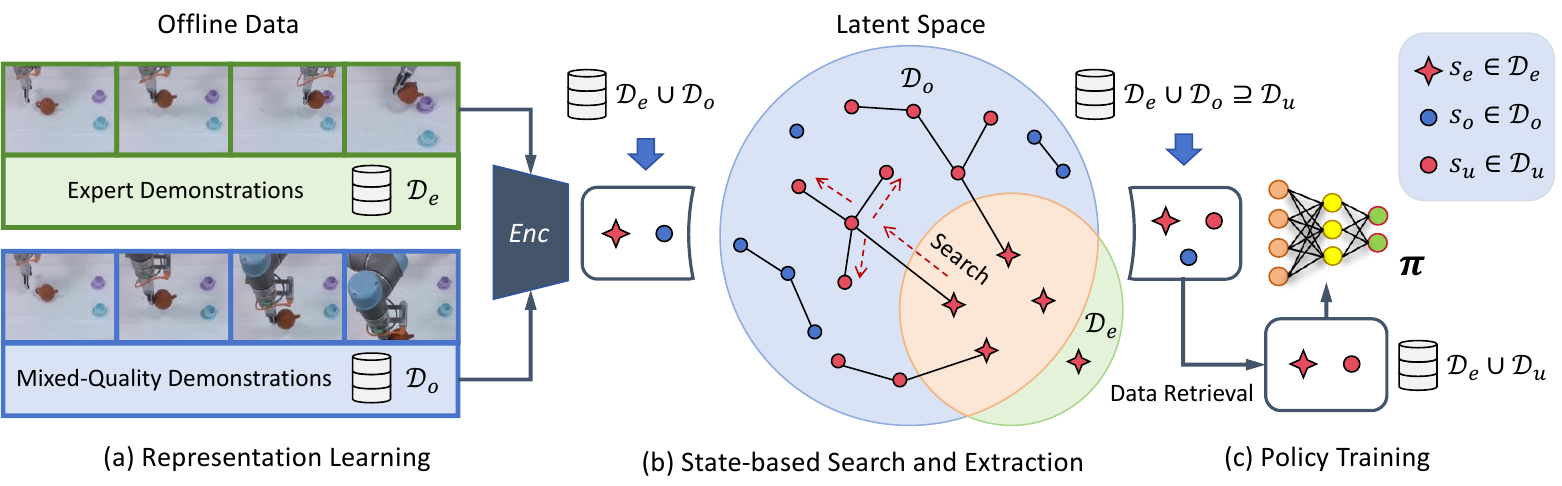} \caption{Overview of the proposed framework: (a) \textit{Representation Learning} utilizes a pre-trained representation model to encode states, establishing a similarity metric based on latent space distances. (b) \textit{State-based Search and Extraction} identifies trajectory segments reflecting beneficial behavior through state-based search, augmenting the training set. (c) \textit{Policy Training} trains an IL policy on the union of expert and retrieved data.}
\label{overall} 
\end{figure*}

In this section, we present our methodology for learning policies from datasets of varying quality in offline IL. Our framework consists of several key stages, outlined as follows: First, we formally define the problem in Section \ref{sec:pf}. Next, we introduce a state similarity metric based on a well-trained encoder in Section \ref{sec:sm}. This is followed by the state-based search process, described in Section \ref{sec:sr}, which retrieves and stitches together useful trajectory segments from imperfect data. Finally, in Section \ref{sec:bc}, we detail the policy training phase, where we train the IL policy on the union of expert and retrieved data. A schematic representation of the framework is shown in Figure \ref{overall}.


\subsection{Problem Formulation}
\label{sec:pf}
We model the problem as a fully observable Markov decision process (MDP), defined by the tuple $ \{\mathcal{S}, \mathcal{A}, P, r, \gamma, d_0\}$, where $\mathcal{S}$ is the state space, $\mathcal{A}$ is the action space, $P: \mathcal{S} \times \mathcal{A} \rightarrow \Delta(\mathcal{S})$ is the transition probability, $r$ is the reward function, $\gamma \in [0,1)$ is the discount factor, and $d_0: \mathcal{S} \rightarrow \Delta(\mathcal{S})$ is the initial state distribution. The agent observes the current state $s_t \in \mathcal{S}$ at each time $t$ and takes an action $a_t \in \mathcal{A}$ according to some policy $\pi(a_t, s_t)$, receiving a reward $r(s_t, a_t)$ and transitioning to the next state $s_{t+1}$ according to $P(s_{t+1}|s_t, a_t)$. The agent's objective is to maximize the cumulative discounted reward:
\begin{equation}
    J(\pi)=\mathbb{E}_{{s_0 \sim d_0,\atop
    s_{t+1} \sim T(\cdot | s_t, \pi(s_t))}}\bigg[\sum_{t=0}^{\infty}\gamma^tr(s_t, a_t)\bigg].\label{eq:accrew}
\end{equation}

In this setup, the true reward function is unknown. The agent has access to offline expert demonstrations $\mathcal{D}_e$, consisting of trajectories ${\{(s_i,a_i,s_{i+1}) |  a_i \sim \pi^e(\cdot|s_i) \}}^{N_e}_{i=0}$ collected using an expert policy $\pi^e$, and a larger set of suboptimal, unlabeled data $\mathcal{D}_o$, which include trajectories ${\{(s_i,a_i,s_{i+1}) | a_i \sim \pi^o(\cdot|s_i) \}}^{N_o}_{i=0}$ obtained by using non-expert policies $\pi^o$. The goal is to learn an optimal policy judiciously utilizing both expert and suboptimal data, aiming to outperform policies learned solely from expert data.

\subsection{State Similarity Metric}\label{sec:sm}
To retrieve valuable behaviors from the suboptimal demonstrations $\mathcal{D}_o$, we introduce a novel similarity metric to compare states. Previous methods such as DWBC \cite{DWBC} and ILID \cite{ILID} use discriminators to compare expert and non-expert states, while behavior retrieval techniques \cite{BehaviorRetrieval} employ variational autoencoders (VAEs) \cite{VAE} to compute the similarity value between states in a latent space. However, these approaches fail to model temporal dependencies, which are essential in the offline IL setting, where maintaining causal relationships in the latent space is critical.

In contrast, we adopt a world model approach \cite{Dreamerv3, zhang2024storm, TDMPC2} to learn state embeddings by predicting multi-step state transitions in a latent space, thereby capturing the temporal dynamics inherent in sequential decision-making tasks. Specifically, we use a simplified world model, which excludes the reward and episode termination prediction, as these are unnecessary for imitation learning. The world model consists of three main components: an encoder, a dynamics predictor, and a decoder:
\begin{equation} \label{eq:comp}
\begin{aligned}
    &\text{Encoder:} &&\quad z_t = q_{\phi}(s_t),\\
    &\text{Dynamics predictor:} &&\quad \hat{z}_t = d_{\phi}(z_{t-1},a_{t-1}),\\ 
    &\text{Decoder:} &&\quad \hat{s}_t = p_{\phi}(z_t),
\end{aligned}
\end{equation}
where $s_t$ and $a_t$ are the states and actions at time $t$, and $z_t$ is the latent representation of state $s_t$. The encoder $q_\phi$ and decoder $p_\phi$ are jointly optimized by minimizing the following loss:
\begin{align}    
         \mathcal{L}(\phi) \!&=\! \mathbb{E}_{(s_t,a_t,s_{t+\!1})_{0:H} \sim {\mathcal{D}_o\cup \mathcal{D}_e}}\!\biggl[\sum_{t=0}^{H} \!\lambda^t({||\hat{z}_{t+\!1}\!-\!{\rm sg}\!\left(q_{\phi}(s_{t+\!1})\right)||}_2^2 \nonumber\\
    & \quad + {||s_t-p_{\phi}(z_t)||}_2^2)\biggl], \label{loss1}
\end{align}
where $\lambda \in (0,1]$ is a decay factor, $H$ is the prediction horizon, and ${\rm sg}$ denotes the stop-gradient operator. The similarity $S(s_i,s_j)$ between any two states $s_i$ and $s_j$ is defined by the $\ell_2$ distance in the latent $z$ space encoded by the world model:
\begin{equation}\label{eq:sim}
    S(s_i, s_j)=-||z_i-z_j||_2=-||q_\phi(s_i)-q_\phi(s_j)||_2.
\end{equation}

\subsection{State-based Search and Extraction}\label{sec:sr}
Using the similarity metric, we perform state-based search to retrieve useful behaviors. Rather than searching at the state-action pair level \cite{DWBC,DGIL,BehaviorRetrieval}, we focus on the state level, enabling more flexible state-action pair retrieval. By starting from expert states, we can trace backward through suboptimal trajectories and stitch together synthetic trajectories that transition to expert states, effectively guiding the agent to recover expert-like behavior during deviations from the expert state distribution. 

The retrieval process begins by comparing states in the suboptimal dataset with those in the expert dataset to identify similar states. We define a selection criterion for a state $s_0\in\mathcal{D}_0$ as follows:
\begin{equation*}
    \mathcal{F}_{\mathcal{D}_e}(s_o) = \frac{\max_{s_e\in \mathcal{D}_e} S(s_e, s_o) - S^-}{S^+ - S^-},
\end{equation*}
where $S^+:=\max_{s_o\in \mathcal{D}_o}\max_{s_e\in \mathcal{D}_e}S(s_e,s_o)$ and $S^-:=\min_{s_o\in \mathcal{D}_o}\max_{s_e\in \mathcal{D}_e}S(s_e,s_o)$ represent the maximum and minimum similarities, respectively, between states in the expert and suboptimal datasets, for normalization. If $\mathcal{F}_{\mathcal{D}_e}(s_o)>\delta$ for a threshold $\delta$, then state $s_0$ is considered similar to an expert state.  The retrieval process is iterated across all states in $s_0\in\mathcal{D}_0$, progressively expanding the set of useful data.

\begin{figure}[htbp] 
\centerline{\includegraphics[width=88mm]{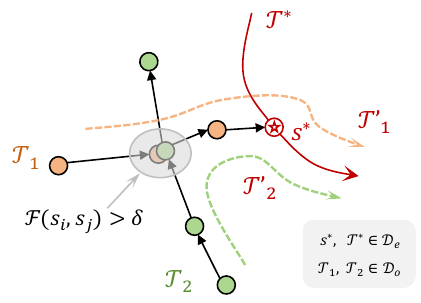}}
\caption{Illustration of the proposed state-based search.}
\label{retrieval}
\end{figure}

As shown in Figure \ref{retrieval}, the state-based search process involves identifying suboptimal trajectories, such as $\mathcal{T}_1$ and $\mathcal{T}_2$, and identifying segments of these trajectories that are similar to expert trajectories, such as $\mathcal{T}^*$. By performing backward tracing from expert states, we can retrieve segments from $\mathcal{T}_1$ and $\mathcal{T}_2$, combining relevant portions of both to generate more informative trajectories, $\mathcal{T}'_1$ and $\mathcal{T}'_2$. This method enables the retrieval of both directly expert-related trajectories as well as segments of suboptimal data that can be stitched to generate informative trajectories.


The steps of the retrieval process are as follows. Let $\mathcal{D}_u^k$ denote the useful data retrieved during step $k$, initialized as $\mathcal{D}_u^k = \emptyset$. For each step $k\in\{1,2,\ldots,K\}$, the useful data is updated based on the criteria described above, progressively enhancing the dataset. For each $s_t^o \in \mathcal{D}_o^0:=\mathcal{D}_o$, if $\mathcal{F}_{\mathcal{D}_e}(s_t^o) > \delta$ and $t - 1 \geq 0$, we add $s_{t-1}^o$ to $\mathcal{D}_u^0$:
\begin{equation} \label{eq:add}
\mathcal{D}_u^0 \gets \left\{ s_{t-1}^o \big| \mathcal{F}_{\mathcal{D}_e}(s_t^o) > \delta, \, t-1 \geq 0, \, s_t^o \in \mathcal{D}_o \right\}.
\end{equation}

This step retrieves transition tuples from non-expert states to expert states, guiding the agent back toward the expert state distribution when it diverges. For each $s_u \in \mathcal{D}_u^0$, if $\mathcal{F}_{\mathcal{D}_e}(s_u) > \delta$, we remove $s_u$ from $\mathcal{D}_u^0$:
\begin{equation} \label{eq:remove}
\mathcal{D}_u^0 \gets \mathcal{D}_u^0 \setminus \left\{ s_u \big| \mathcal{F}_{\mathcal{D}_e}(s_u) > \delta, \, s_u \in \mathcal{D}_u^0 \right\}.
\end{equation} 
Equation \eqref{eq:remove} ensures that the selected transition tuples do not solely come from expert-to-expert states, thus avoiding deviation or cyclic behavior in expert states.

For subsequent retrieval steps, let $K$ represent the total number of search iterations. During search step $k+1$, we update $\mathcal{D}_u^k$ as follows:
\begin{align} \label{eq:add_i}
\mathcal{D}_u^k & \gets \left\{ s_{t-1}^o \big| \mathcal{F}_{D^k_e}(s_t^o) \!> \!\delta, \, t-\!1 \geq \!0, \, s_t^o \in\! D^k_o  \right\},\\
D^k_e & \gets \mathcal{D}_e \cup \bigcup_{j=0}^{k-1} \mathcal{D}_u^j, \quad  D^k_o \gets \mathcal{D}_o \setminus \bigcup_{j=0}^{k-1} \mathcal{D}_u^j,
\end{align}
and remove states from $\mathcal{D}_u^k$ where the similarity with expert states exceeds the threshold: 
\begin{equation} \label{eq:remove_i}
\mathcal{D}_u^k \gets \mathcal{D}_u^k \setminus \left\{ s_u \big| \mathcal{F}_{D^k_e}(s_u) > \delta, \, s_u \in \mathcal{D}_u^k \right\}.
\end{equation}

By iterating backward from expert states and performing state-based retrieval across imperfect data, we generate synthetic trajectories leading to expert states. Finally, the retrieved data is:
\begin{equation*}
    \mathcal{D}_u = \bigcup_{k=0}^{K-1} \mathcal{D}_u^k,
\end{equation*}
which contains diverse behavioral patterns to guide the agent beyond the expert state distribution.

\subsection{Behavior Cloning}\label{sec:bc}
Once the useful dataset $\mathcal{D}_u$ has been constructed, we proceed to train the agent using both the expert as well as the retrieved data with a behavior cloning loss. The objective is to jointly learn the policy $\pi_{\theta}(a|s)$ from the expert data $\mathcal{D}_e$ and the selected useful data $\mathcal{D}_u$, which is formalized as follows:
\begin{equation}
\min_{\theta} \mathbb{E}_{(s,a) \sim \mathcal{D}_e}[-\log\pi_{\theta}(a|s)]+\mathbb{E}_{(s,a) \sim \mathcal{D}_u}[-\log\pi_{\theta}(a|s)].\label{loss2}
\end{equation}

As the number of search steps increases, the retrieved states may increasingly deviate from the expert states. Therefore, we introduce a decay factor for the data in 
$\mathcal{D}_u$ during the policy learning process to ensure stable convergence.

\begin{algorithm}[H]
\caption{SBR: State-based Retrieval and Policy Learning}
\begin{algorithmic}[1]
\STATE \textbf{Input:} Expert data $\mathcal{D}_e$, suboptimal data $\mathcal{D}_o$
\STATE Initialize $\phi$, $\theta$ randomly, $\mathcal{D}_u \gets \emptyset$, decay factor $\gamma$, search steps $K$
\STATE \label{line:rep-learning-start} // Representation learning
\WHILE{not converged} 
\STATE Sample $(s_t, a_t, s_{t+1}) \sim \mathcal{D}_e \cup \mathcal{D}_o$
\STATE Update $\phi$ using \eqref{loss1}
\ENDWHILE \label{line:rep-learning-end}
\STATE \label{line:retrieval-start} 
// State-based search and behavior retrieval
\FOR{$k = 0$ to $K-1$}
\STATE Build $\mathcal{D}_u$ using \eqref{eq:add_i} and \eqref{eq:remove_i}
\ENDFOR \label{line:retrieval-end}
\STATE \label{line:policy-learning-start} // Policy learning
\STATE \label{line:policy-learning-end} Update $\theta$ using \eqref{loss2}.
\end{algorithmic}\label{alg:sbr}
\end{algorithm}

The pseudocode for our algorithm is outlined in Algorithm \ref{alg:sbr}, where lines \ref{line:rep-learning-start}--\ref{line:rep-learning-end} describe the representation learning phase, lines \ref{line:retrieval-start}--\ref{line:retrieval-end} summarize the state-based retrieval process, and lines \ref{line:policy-learning-start}--\ref{line:policy-learning-end} describe the policy learning phase.

\begin{figure*}[t]
    \centering
    \includegraphics[width=174mm]{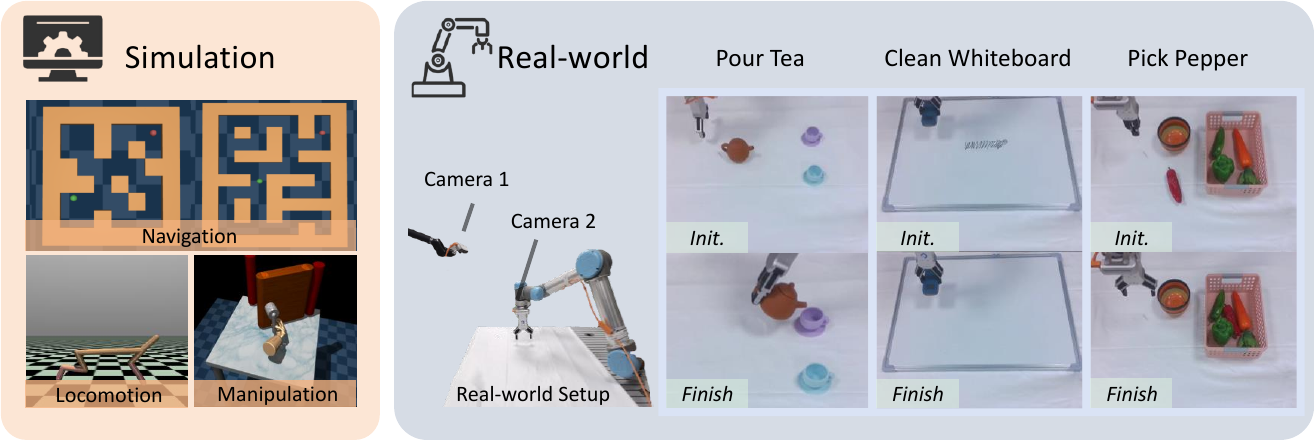} 
    \caption{Illustration of the tasks used in the experiments. Left: Tasks from the D4RL benchmark. Right: Our real-world tasks.}
    \label{env}
\end{figure*}

\section{Experiments}\label{sec:exp}
In this section, we evaluate the proposed method in both simulation and real-world environments to assess its effectiveness across various tasks, as shown in Figure \ref{env}. First, we compare the performance of our method with state-of-the-art IL baselines in simulation benchmarks. We then conduct both qualitative and quantitative analyses to demonstrate how our method can effectively retrieve useful behaviors from imperfect data, enabling the agent to learn a robust policy that outperforms existing methods.

\begin{table*}[t] 
\centering
\caption{Performance comparison on the D4RL benchmark. All results are normalized between $0$ and $100$, where $0$ indicates random policy performance, and $100$ corresponds to expert-level performance \cite{D4RL}. The best performance in each category is highlighted in bold, while second- and third-place results are underlined.}
\label{tab:performance_comparison}
\setlength{\tabcolsep}{3pt}
\begin{tabular}{c|c|cccccccc}
\toprule
Domain & DataSet & BC-exp & BC-all & DemoDICE & DWBC & UDS & ILID & SBR (Ours)\\
\midrule
\multirow{3}{*}{Navigation} & sparse-umaze-v1 & $88.9\pm42.0$ & $47.1\pm13.0$ & $15.7\pm1.66$ & $\underline{128.\pm14.5}$ & $91.1\pm22.9$ & $\mathbf{135.\pm3.15}$ & $\underline{104.\pm9.46}$\\ 
& sparse-medium-v1 & $38.3\pm18.1$ & $5.55\pm7.89$ & $24.4\pm7.63$ & $80.4\pm16.4$ & $\underline{97.0\pm20.0}$ & $\underline{114.\pm33.3}$ & $\mathbf{152.\pm3.15}$\\
& sparse-large-v1 & $1.45\pm6.63$ & $23.7\pm21.4$ & $60.7\pm30.6$ & $\mathbf{161.\pm43.7}$ & $\underline{134.\pm26.0}$ & $12.0\pm14.5$ & $\underline{83.6\pm10.2}$\\
\midrule
\multirow{4}{*}{Locomotion} & hopper-medium-v2 & $\underline{72.9\pm5.50}$ & $56.4\pm1.86$ & $54.1\pm1.67$ & $\underline{88.1\pm4.71}$ & $66.0\pm0.49$ & $46.7\pm33.4$ & $\mathbf{94.24\pm4.87}$\\ 
& halfcheetah-medium-v2 & $13.3\pm2.74$ & $\underline{42.8\pm0.41}$ & $41.1\pm1.00$ & $22.5\pm3.94$ & $\mathbf{57.1\pm6.91}$ & $40.9\pm1.71$ & $\underline{47.9\pm4.50}$\\
& walker2d-medium-v2 & $\underline{99.1\pm3.66}$ & $\underline{86.8\pm5.28}$ & $73.0\pm2.09$ & $84.8\pm5.65$ & $8.52\pm4.99$ & $85.5\pm2.66$ & $\mathbf{102.\pm7.64}$\\
& ant-medium-v2 & $51.3\pm6.87$ & $\mathbf{98.7\pm3.68}$ & $\underline{91.2\pm3.79}$ & $37.5\pm5.95$ & $18.4\pm10.5$ & $\underline{98.3\pm18.8}$ & $54.4\pm1.64$\\
\midrule
\multirow{4}{*}{Manipulation} & pen-cloned-v1 & $\underline{68.6\pm35.0}$ & $5.89\pm8.01$ & $33.1\pm10.9$ & $\underline{75.6\pm27.2}$ & $4.32\pm8.03$ & $45.2\pm6.05$ & $\mathbf{95.6\pm6.28}$\\
& door-cloned-v1 & $\underline{5.25\pm7.90}$ & $0.02\pm0.04$ & $0.07\pm0.09$ & $0.36\pm0.33$ & $-.33\pm0.01$ & $\underline{8.28\pm7.82}$ & $\mathbf{25.6\pm15.9}$ \\
& hammer-cloned-v1 & $\underline{101.\pm17.7}$ & $0.28\pm0.00$ & $0.24\pm0.01$ & $\underline{98.6\pm8.40}$ & $0.38\pm0.07$ & $80.6\pm32.6$ & $\mathbf{119.\pm0.90}$ \\
& relocate-cloned-v1 & $\underline{59.8\pm32.9}$ & $10.5\pm5.23$ & $-0.1\pm0.09$ & $56.2\pm23.7$ & $-.32\pm0.03$ & $\underline{61.6\pm3.57}$ & $\mathbf{69.9\pm6.38}$ \\
\bottomrule
\end{tabular}
\end{table*}

\subsection{Experiment Setup}

\subsubsection{Simulation Environments} 
We evaluate our approach using the D4RL benchmark, which provides a series of tasks across three domains: navigation, locomotion, and manipulation. These tasks serve as a comprehensive testbed for assessing the generalizability and performance of the proposed method across a wide range of robotic control challenges, which have been widely used in previous works \cite{DemoDICE,DWBC,UDS,ILID}.

\begin{itemize}
    \item \textbf{Navigation.} 
    In the Maze2D environment, the agent is required to navigate through a maze to reach a fixed target goal and stay there. The D4RL benchmark includes three maze layouts (i.e., umaze, medium, and large). We utilize five expert trajectories generated by a path planner \cite{D4RL} as the expert dataset and consider the offline data as $1,000$ logged experiences with random goals. 
    \item \textbf{Locomotion.} In the Locomotion environment, the agent is required to achieve the desired motion patterns through joint control. It consists of four different environments (i.e., hopper, walker2d, halfcheetah, and ant). We use five expert trajectories from the ``-expert'' dataset and consider the offline data as $1,000$ trajectories from the ``-medium'' dataset, which comes from an early-stopped SAC policy.
    \item \textbf{Manipulation.} In the Manipulation environment, the agent is required to perform complex manipulation tasks. It consists of four environments (i.e., hammer, door, pen, and relocate), where a simulated $24$-DoF Shadow Hand robot is controlled to perform tasks such as hammering a nail and opening a door. We use $50$ expert trajectories from the ``-expert'' dataset and consider the offline data as $1,000$ trajectories from the ``-cloned'' dataset, which is generated by an imitation policy.
\end{itemize}

\subsubsection{Real-world Environments} 
In the real world, we design tasks of varying complexities to comprehensively evaluate our method.

\begin{itemize}
    \item \textbf{Pour Tea.}  In this task, the UR5 robot is required to grasp a teapot on the table, position it above the target cup, and tilt it at a specific angle to pour water into the cup.

    \item \textbf{Clean Whiteboard.} In this task, the UR5 robot uses an eraser to wipe marker stains from a whiteboard. Successful removal of the stains is considered task completion.
    
    \item \textbf{Pick Pepper.} In this task, the UR5 robot is required to pick a pepper from the table and place it into a basket. 

\end{itemize}

For real-world data collection, we deploy a teleoperation system where the operator controls the robot via a 3Dconnexion SpaceMouse at $10$Hz during demonstrations. For each task, we collect only $20$ demonstrations as expert data, which is significantly fewer than in previous works \cite{DiffusionPolicy,ACT3D,Aloha,QATT}. The imperfect demonstrations including image observations come from approximately $100$ failed trajectories during the collection process, as well as trajectories with different subtask goals or random arm movements. The UR5 robot station is equipped with two RealSense D435 depth cameras, which are downsampled to a 256x256 resolution and used for policy input. Additionally, the policy incorporates the robot arm’s proprioception as input. For all tasks, we employ position control, with the policy directly outputting the desired end-effector pose.

\subsection{Baselines}
In our experiments, we compare the proposed method against the following baselines:
\begin{itemize}
    \item \textbf{BC-exp:} Behavioral cloning (BC) on the expert data $\mathcal{D}_e$, which is limited in quantity and prone to significant compounding errors due to distribution shift.
    \item \textbf{BC-all:} Behavioral cloning on all available data $\mathcal{D}_e \cup \mathcal{D}_o$. Given that $\mathcal{D}_o$ contains a significant portion of low-quality data, this approach negatively impacts the policy and leads to suboptimal performance. 
    \item \textbf{DemoDICE:} DemoDICE utilizes offline data by applying a constraint over $\mathcal{D}_o$ to ensure proper policy regularization, while simultaneously applying a similar regularization to $\mathcal{D}_e$ for expert imitation.
    \item \textbf{DWBC:} DWBC incorporates offline data by training a discriminator through positive-unlabeled learning, which re-weights the behavioral cloning objective for improved policy learning. 
    \item \textbf{UDS:} UDS assigns a reward label of zero to data in $\mathcal{D}_o$ and a reward label of one to data in $\mathcal{D}_e$. The policy is trained using offline reinforcement learning on the merged dataset.
    \item \textbf{ILID:} ILID identifies positive behaviors based on the
     resultant states and retrieves beneficial behaviors from imperfect demonstrations. It then applies behavioral cloning on both the expert and the retrieved data for policy improvement.
    
\end{itemize}

For simulation tasks, we use a three-layer perceptron as the policy network. In real-world robotic tasks, we employ BAKU \cite{BAKU}, a transformer-based policy network. Additionally, to calculate the similarity between pixel states, we incorporate R3M \cite{R3M}, a pre-trained feature representation for robotic manipulation tasks.

\subsection{Simulated Results} 
We evaluate the proposed method on $11$ tasks from the D4RL benchmark. As shown in Table \ref{tab:performance_comparison}, our method consistently outperforms the baselines on $7$ out of $11$ tasks, while remaining highly competitive in the remaining tasks. For the navigation domain, we excluded the encoder due to the low-dimensional observation space. To ensure a fair comparison, we refrained from initializing the policy parameters with behavioral cloning during training.

The following observations can be made from the results:
\begin{itemize}
    \item \textbf{BC-exp}: The agent struggles to learn a satisfactory policy with limited expert data, as seen in the poor performance of BC-exp.
    \item \textbf{BC-all}: Directly applying behavioral cloning to the entire dataset, including suboptimal data, results in suboptimal performance, as indicated by BC-all's poor performance.
    \item \textbf{DemoDICE and DWBC}: These methods perform well when the quality of the offline data is high (e.g., in the \emph{walker} and \emph{ant} settings). However, their performance significantly deteriorates when the offline data quality is low.
    \item \textbf{UDS}: The reinforcement learning approach employed by UDS enables the agent to implicitly learn transitions from non-expert states to expert states by optimizing long-term reward returns. However, the reinforcement learning paradigm is prone to instability due to the ``deadly triad" issue, which can hinder learning.
\item \textbf{ILID}: ILID trains a discriminator similar to DWBC to differentiate between expert and non-expert states and selects transition segments from non-expert to expert states. Our method shares similar insights with ILID but achieves significant performance improvements by leveraging beneficial behavioral patterns contained in stitched state segments.
\end{itemize}

In conclusion, our method effectively utilizes imperfect data, offering a viable approach to mitigating the covariate shift problem in IL. By learning from both expert and non-expert data, our approach provides a more robust and flexible solution for training policies in challenging environments.

\subsection{Ablation Study}
The encoder plays a pivotal role in our method by learning environmental dynamics from offline data and capturing state similarities in the encoded space. To better understand the impact of the encoder, we conduct ablation studies on manipulation tasks, evaluating the performance with no encoder or using a simple autoencoder for encoding.
 
\begin{table}[t] 
\centering
\caption{Ablation study with encoder.}
\label{tab:Ablation}
\setlength{\tabcolsep}{5pt}
\begin{tabular}{cccc}
\toprule
Dataset  & w/o encoder & autoencoder & world model \\
\midrule
pen-cloned-v1  & $66.8\pm6.49$ & $93.5\pm1.15$ & $\mathbf{95.6\pm6.28}$\\
door-cloned-v1  & $3.77\pm3.74$ & $7.04\pm3.66$ & $\mathbf{25.6\pm15.9}$ \\
hammer-cloned-v1  & $93.3\pm20.2$ & $112.\pm5.07$ & $\mathbf{119.\pm0.90}$ \\
relocate-cloned-v1  & $13.1\pm6.97$ & $59.2\pm5.14$ & $\mathbf{69.9\pm6.38}$ \\
\bottomrule
\end{tabular}
\end{table}

\begin{table}[t] 
\centering
\caption{Comparison under varying numbers of expert trajectories.}
\label{tab:Ablation2}
\setlength{\tabcolsep}{4.5pt}
\begin{tabular}{ccccc}
\toprule
Dataset & \multicolumn{2}{c}{pen-cloned-v1} & \multicolumn{2}{c}{hammer-cloned-v1} \\
\cmidrule(lr){2-3} \cmidrule(lr){4-5}
$N_e$ & BC-exp & SBR & BC-exp & SBR \\
\midrule
10 & $72.1\pm22.2$ & $\mathbf{75.9\pm1.31}$ & $5.96\pm4.56$ & $\mathbf{12.9\pm1.33}$ \\
20 & $64.6\pm12.7$ & $\mathbf{86.3\pm0.34}$ & $72.4\pm4.82$ & $\mathbf{85.2\pm1.44}$ \\
30 & $\mathbf{87.5\pm0.85}$ & $76.1\pm1.00$ & $52.8\pm9.51$ & $\mathbf{114.\pm6.55}$ \\
40 & $83.4\pm2.85$ & $\mathbf{96.6\pm10.2}$ & $95.2\pm17.2$ & $\mathbf{119.\pm0.22}$ \\
50 & $68.6\pm35.0$ & $\mathbf{95.6\pm6.28}$ & $101.\pm17.7$ & $\mathbf{119.\pm0.90}$ \\
\bottomrule
\end{tabular}
\end{table}

\begin{figure*}[t]
\centerline{\includegraphics[width=175mm]{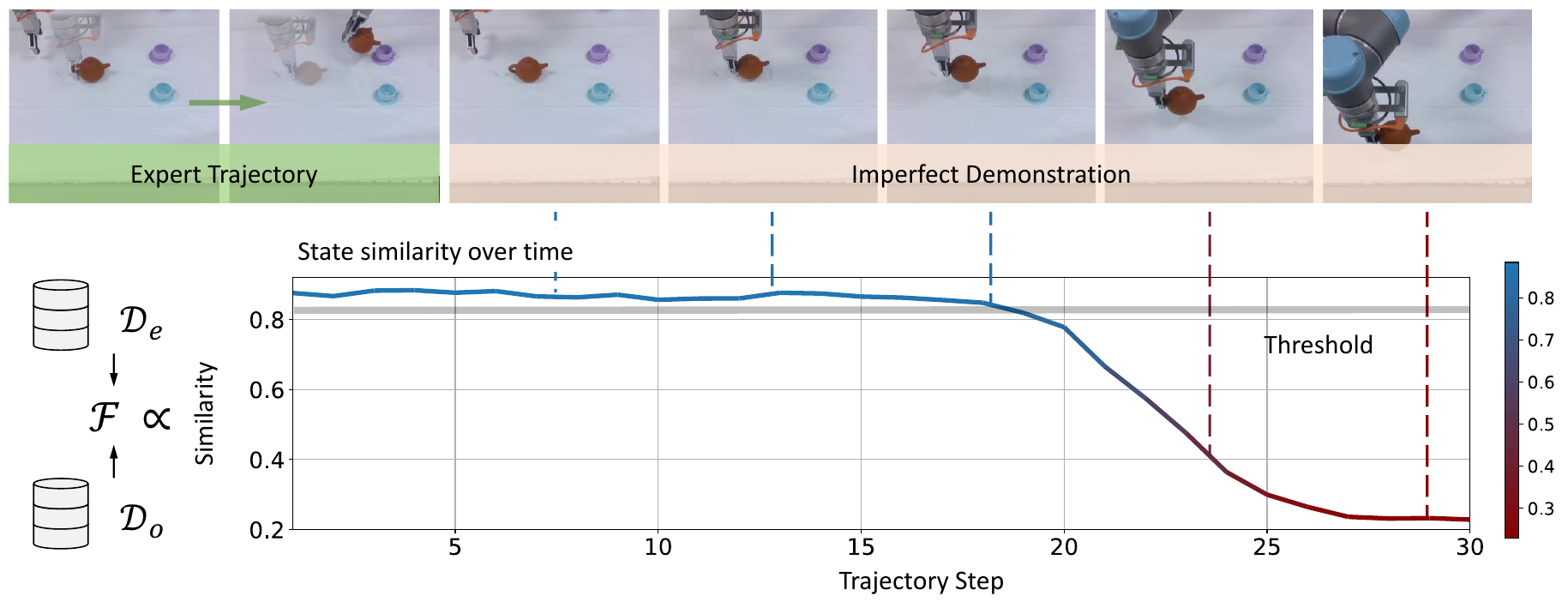}}
\caption{Demonstration of behavior retrieval by SBR. The similarity between each state in imperfect demonstrations and the expert trajectory is calculated and compared with the threshold.}
\label{dis}
\end{figure*}

\begin{figure}[t] 
\centerline{\includegraphics[width=86mm]{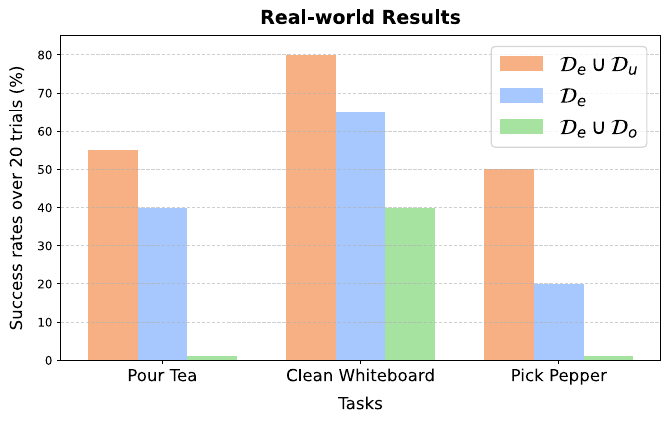}}
\caption{Results of real-world experiments. The policy success rates averaged over $20$ trials for each setup are reported.}
\label{bar}
\end{figure}

Table \ref{tab:Ablation} shows that omitting the encoder and directly performing similarity computation and policy training on raw states leads to suboptimal performance. Using a simple autoencoder improves performance but still lags behind the performance with the proposed simplified world model. This demonstrates the benefit of incorporating the prediction module, which enables the model to learn shared environmental dynamics with expert data from imperfect demonstrations, optimizing the encoding space.

To investigate the effect of varying the number of expert trajectories on policy performance, we trained policies using $10$ to $50$ expert trajectories while keeping the size of suboptimal data fixed in the \emph{pen} and \emph{hammer} environments. Table \ref{tab:Ablation2} presents the comparison between SBR and behavioral cloning across different settings, showing that SBR remains competitive even when expert data is scarce.

\subsection{Real-world Results}
In real-world robotic experiments, we first validate our similarity criterion function by analyzing the behavioral patterns retrieved from imperfect demonstrations. Figure \ref{dis} illustrates the similarity between suboptimal and expert trajectories. The robot initially attempts to grasp the teapot, exhibiting a behavior pattern consistent with expert demonstrations (high similarity region). As the trajectory progresses, the robot fails to position the teapot above the teacup (low similarity region). The states with high similarity are then selected as starting points for backward tracing to extract useful trajectory segments.

The real-world robotic experiments demonstrate that our method effectively utilizes imperfect demonstration data, leading to significant improvements in policy performance and generalization. The results, shown in Figure \ref{bar}, indicate that our method achieves performance improvements exceeding $23\%$ across downstream tasks.

\section{Conclusions} 
In this paper, we propose state-based retrieval (SBR), a novel offline IL method that effectively leverages imperfect, unlabeled data by learning shared environmental dynamics and retrieving beneficial behavioral patterns from stitched trajectories. The state-based retrieval method enables us to stitch trajectory segments from imperfect data and form new ``expert trajectories,'' while the well-designed encoder facilitates the learning of environmental dynamics across mixed-quality datasets. Although our approach demonstrates a comprehensive utilization of offline data, certain limitations remain. A notable limitation is the reliance on the state similarity measures represented by distances in the latent space. When the encoder is not well-trained or insensitive to task-relevant features (e.g., in pixel-based inputs), the retrieval process may introduce errors and lead to inaccurate state selections. In future work, we aim to explore more effective similarity metrics to address these challenges.


\bibliographystyle{IEEEtran}
\bibliography{ieeeconf/main}

\end{document}